\documentclass[conference]{IEEEtran}
\IEEEoverridecommandlockouts
\usepackage{cite}
\usepackage{amsmath,amssymb,amsfonts}
\usepackage{algorithmic}
\usepackage{graphicx}
\usepackage{textcomp}
\usepackage{xcolor}
\usepackage{amsmath}
\usepackage[bookmarks=false]{hyperref} 
\usepackage{colortbl}
\usepackage{booktabs}
\usepackage[per-mode=symbol,range-phrase=--,range-units=single]{siunitx}
\usepackage{nicefrac}
\usepackage{multirow}
\usepackage{calc}
\usepackage{printlen}
\usepackage[normalem]{ulem}
\usepackage{ragged2e}
\usepackage{amsmath}
\usepackage{graphicx}
\usepackage{amssymb}
\usepackage{multirow}
\usepackage{makecell}
\usepackage{mathrsfs}
\usepackage{multirow}
\usepackage[normalem]{ulem}
\useunder{\uline}{\ul}{}
\usepackage{threeparttable}
\usepackage{color}
\usepackage{booktabs}
\usepackage{algorithm}
\usepackage{algorithmic}
\def\BibTeX{{\rm B\kern-.05em{\sc i\kern-.025em b}\kern-.08em
   T\kern-.1667em\lower.7ex\hbox{E}\kern-.125emX}}

\usepackage{booktabs}

\usepackage{bbm}

\begin{document}

\title{Multi-level Graph Subspace Contrastive Learning for Hyperspectral Image Clustering}

\author{\IEEEauthorblockN{Jingxin Wang$^{\dag}$$^{1}$, Renxiang Guan$^{\dag}$$^{2}$, Kainan Gao$^{1}$, Zihao Li$^{1}$, Hao Li$^{2}$,  Xianju Li$^{\ast}$$^{1}$ , Chang Tang$^{1}$}
\thanks{$^{\dag}$ Equally contributed. *Corresponding author}
\IEEEauthorblockA{\textsuperscript{\rm 1}\textit{School of Computer Science, China University of Geosciences, Wuhan, 430074, China}\\
\textsuperscript{\rm 2}\textit{School of Computer, National University of Defense Technology, Changsha, 410073, China} \\
}
renxiangguan@nudt.edu.cn; ddwhlxj@cug.edu.cn\\
}

\maketitle

\begin{abstract}
\looseness=-1 Hyperspectral image (HSI) clustering is a challenging task due to its high complexity. Despite subspace clustering shows impressive performance for HSI, traditional methods tend to ignore the global-local interaction in HSI data. In this study, we proposed a multi-level graph subspace contrastive learning (MLGSC) for HSI clustering. The model is divided into the following main parts. Graph convolution subspace construction: utilizing HSI’s spectral and texture feautures to construct two graph convolution views. Local-global graph representation: local graph representations were obtained by step-by-step convolutions and a more representative global graph representation was obtained using an attention-based pooling strategy. Multi-level graph subspace contrastive learning: multi-level contrastive learning was conducted to obtain local-global joint graph representations, to improve the consistency of the positive samples between views, and to obtain more robust graph embeddings. Specifically, graph-level contrastive learning is used to better learn global representations of HSI data. Node-level intra-view and inter-view contrastive learning is designed to learn joint representations of local regions of HSI. The proposed model is evaluated on four popular HSI datasets: Indian Pines, Pavia University, Houston, and Xu Zhou. The overall accuracies are 97.75\%, 99.96\%, 92.28\%, and 95.73\%, which significantly outperforms the current state-of-the-art clustering methods.

\end{abstract}


%
%

\section{Introduction}
Owing to the unique imaging mechanism of hyperspectral imagery (HSI), which contains a large amount of spatial and spectral information \cite{ResCapsNet}, it has been widely used in various fields such as thematic mapping of vegetation \cite{2yang2022enhanced}, geological exploration \cite{guan4}, medical imaging, resource management 
\cite{4wang2020identification}, \cite{Guan2}, etc. The clustering task is one of the applications of HSI, which aims to group similar pixels into the same group to discover potential patterns or structures in the data without the need for pre-defined category labels \cite{zhai2021hyperspectral}, \cite{2024AMGC}. Although HSI clustering has been extensively studied by many scholars, it is still a challenging task due to the high dimensionality and complex spatial distribution of land cover types  \cite{5peng}, \cite{wen1}.

In recent years, it has been found that subspace clustering methods excel at the task of dealing with high-dimensional data, showing very reliable performance \cite{6vidal2011subspace}, \cite{wen5}, \cite{9zhang}, \cite{wen3}, \cite{wen4}. The better-performing methods, such as sparse subspace clustering (SSC) \cite{10elhamifar2013sparse}, \cite{wen2}, utilize the sparsity of the data representation to discover the subspace structure in the data by representing the data points as linear combinations of other data points. Zhai et al \cite{35zhai2016new} proposed a new effective $l_2$-paradigm regularized SSC algorithm, which adds a four-neighbourhood $l_2$-paradigm regularizer to the classical SSC model to enhance the segmental smoothness of sparse coefficient matrices and the homogeneity of the final clustering results. However, due to the nonlinear structure of HSI data, traditional subspace clustering methods tend to ignore the potential multiscale structure in the HSI data, which results in the loss of some critical information \cite{wuda1}, \cite{wuda2}, \cite{AMFGCN}. Cai et al \cite{12cai} proposed a new subspace clustering framework (GCSC) for robust HSI clustering, which reconverts the self-expression properties of the data to non-Euclidean domains to produce a more robust graph embedding dictionary \cite{uestc1}, \cite{uestc2}, \cite{uestc3}, \cite{uestc4}, \cite{uestc5}.

While subspace clustering excels in exploring the underlying structural relationships of the data, to further mine and exploit the complex topology of the data, graph convolutional neural networks (GCNs) \cite{15yang2020probabilistic}, \cite{wang2022graph} provide an effective framework for learning deeper node representations by operating directly on graph structures. Realizing that existing methods fail to exploit higher-order relationships and long-term interdependencies, Zhang et al. \cite{9zhang} propose hypergraph convolutional subspace clustering (HGCSC), which converts the classical self-representation into a hypergraph convolutional self-representation model. Realising that existing subspace clustering methods often neglect higher-order feature extraction, Cai et al \cite{17cai2021graph} proposed a graph regularised residual subspace clustering network (GR-RSCNet) that jointly learns deep spectral space representations and robust nonlinear affinities via deep neural networks. Liu et al. \cite{38Liu} propose a novel model defined on graph convolution (GCOT) for HSI spectral clustering to capture the intrinsic geometric structure of HSI \cite{hao1}, \cite{hao2}.

However, GCN still has some limitations in capturing the global structure of graphs and processing large-scale data, so graph contrastive learning has received a lot of attention from scholars. Guan et al. \cite{PSCPC} proposed a pixel-superpixel level contrastive learning and pseudo-label correction method (PSCPC) to efficiently identify and extract fine-grained features in HSI by performing contrastive learning within superpixels. Cai et al. \cite{NCSC} proposed a new neighborhood contrastive regularization method (NCSC) to maximize the consistency between positive samples in the subspace and strengthen the subspace representation at the superpixel level. In order to fully utilize the role of intra-view and inter-view contrastive learning in learning joint representations of local regions of HSI, multi-view clustering methods have attracted great interest and achieved remarkable results \cite{li2023mixture}.

Benefiting from the complementary information of multiple views at multi-view clustering methods have achieved considerable success in various fields. Guan et al. \cite{CMSCGC} proposed a graph convolutional network-based HSI contrastive multi-view subspace clustering method (CMSCGC) that adequately realizes interactions between different views. Liu et al. \cite{22liu2022multilayer} proposed a general and effective self-encoder framework for multi-level graph clustering that utilizes a contrastive fusion module to capture the consistency information between different levels. Li et al \cite{24li2022deep} proposed a Deep Mutual Information Subspace Clustering (DMISC) network that reduces the clustering overlap problem by maximising the mutual information between the samples to extend inter-class dispersion and intra-class tightness. However, existing contrastive learning clustering methods often overlook the use of HSI global information, which may lead to the neglect of the internal structure of the data \cite{hutao1}, \cite{hutao2}, \cite{hutao3}.

To overcome the above difficulties, a HSI subspace clustering framework based on multi-level graph contrastive learning (MLGSC) is proposed in this article, which introduces an attention pooling module after multi-feature correlation extraction to obtain a more representative global graph representation. In addition, a contrastive learning mechanism is introduced into the node-level graph representation and global graph representation, respectively, to fully utilize the local information of HSI features, global information, and complementary information among different views to obtain more robust graph embeddings. Finally, we perform spectral clustering on the constructed affinity matrices to obtain clustering results for HSI. Figure \ref{fig:main} visualizes the motivation of MLGSC. In summary, the main contributions of this work are as follows.

\begin{enumerate}
    \item A subspace clustering framework MLGSC based on multi-level graph contrastive learning is proposed, which introduces a contrastive learning mechanism in the node-level representation and global graph-level representation, respectively, to efficiently extract the key features and utilize the complementary information between different views by obtaining a joint local-global graph representation.
    \item An attention pooling module is proposed to measure the node representations of views and emphasizes the contribution of each node to the global graph representation in order to obtain a more representative global graph-level representation.
    \item Our experimental results on four widely recognized HSI datasets show that the proposed subspace clustering model is more effective than many existing HSI clustering methods. The successful attempt of MLGSC provides ideas for unsupervised clustering of HSI.
\end{enumerate}

The rest of the paper is organized as follows: in Section \ref{se:related_work}, we first briefly review related work on multi-level subspace clustering, contrastive learning strategies, and graph-level representation learning. Section \ref{se:method} presents the details of our proposed MLGSC method. In Section \ref{se:experiments}, we give experimental results and empirical analysis. Finally, we summarize the work in Section \ref{se:conclusion}.

%
%

\section{Related Work}
\label{se:related_work}
\subsection{Multi-view Subspace Clustering}

Single view clustering is often limited by the expressive ability of a single data view when dealing with complex data, which is difficult to cope with the diversity and complexity of HSI data and may lead to skewed or inaccurate clustering results. For HSI data with noisy or incomplete information, multi-view clustering can effectively resist fluctuations in data quality and improve the accuracy and robustness of clustering by combining information from multiple data views. However, multi-view clustering also faces challenges in data integration, feature alignment, and algorithm design, and requires more complex processing flow and algorithm design.

 Let \(X=[x_1,x_2,\ldots,x_N]\in\mathbb{R}^{M\times N}\) be a collection of \(N\) data points extracted from an \(M\)-dimensional space, belonging to \(k\) distinct sets of clusters and lying in the concatenation of linear subspaces \(R_1\cup R_2\cup\ldots\cup R_k\) in which the self-representation assumes that each \(x_i\) can be represented by a linear combination of data points belonging to the same subspace, describing the subspace self-representation model as
\begin{equation}
X = XC + E
\end{equation}
where $X$ is the matrix representation of the data points, $C$ is the self-representation matrix, $E$ is the error matrix, and \(C = [c_1,c_2,\ldots,c_w]\in\mathbb{R}^{N\times N}\), $w$ represents the dimension of the self-representation matrix $C$, matrix \(C\) will be obtained by the following criterion:
\begin{equation}
\min_C \left\|C\right\|_1 \quad \text{s.t.} \quad X=XC+E, \text{diag}(C)=0
\end{equation}

Among them, \(\left\|\cdot\right\|_1\) is the number of \(l_1\)-paradigm numbers, \(\text{diag}(C)=0\) is used to avoid mundane solutions. Formally, multi-view subspace clustering is described as:
\begin{equation}
\begin{gathered}
\arg\min _{C^p} \left\|X^p-X^p C^p\right\|_F^2+\lambda f\left(C^p\right) \\
\text { s.t. }  {diag}(C^p) \geq 0, C^{p\top} \mathbf{1}=\mathbf{1},
\end{gathered}
\end{equation}
where \(\lambda\) is the regularization factor, and \(\left\|C^p\right\|\) is the representation of the \(p\)-th subspace, which is determined by different forms of matrix paradigms, e.g., the kernel paradigm is used for subspace clustering of low-rank representations (LRR), and the sparse subspace clustering uses the \(l_1\)-norm. A large number of multi-view clustering methods have been proposed by many scholars, Kopriva et al \cite{25brbic2018multi} obtain a joint subspace representation of all views by learning affinity matrices subject to sparsity and low-rank constraints \cite{xia1}, \cite{xia2}, \cite{xia3}, \cite{xia4}, \cite{xia5}. These methods are designed for feature matrices and attempt to learn graphs from data. To directly cluster multiple graphs, self-weighted multi-view clustering (SwMC) methods learn shared graphs from multiple graphs by using a new weighting strategy \cite{hao3}.

\subsection{Contrastive Learning}

Contrastive learning aims to learn discriminative representations by comparing positive and negative samples, relying on the establishment of such samples \cite{SCGN}, \cite{Dinknet}, \cite{DFCN}, \cite{DGCN}, \cite{HSAN}. In computer vision, pairs of positive and negative samples can be generated using multi-level enhancement channels. In recent years, contrastive learning has been introduced to remote sensing image processing tasks\cite{44}. Inspired by SimCLR, we use contrastive learning to train an encoder for hyperspectral following semantic segmentation network.

\textbf{Contrastive Loss.} Contrastive loss expects positive sample pairs to be similar and negative sample pairs to be dissimilar. Specifically, a sample of number \(N\) is expanded to \(2N\) samples, a pair of samples expanded from the same sample constitutes a positive pair, and the other \(2(N-1)\) samples are negative samples. Therefore, the contrastive loss \(L_Con\) is defined as:
\begin{equation}\mathcal{L}_C=\frac1{2N}\Sigma_{k=1}^N(\ell(\tilde{x}_i,\hat{x}_i)+\ell(\hat{x}_i,\tilde{x}_i))\end{equation}
\begin{equation}\ell(\tilde{x}_{i},\hat{x}_{i})=-log\frac{\exp\left(\sin\left(\tilde{z}_{i},\hat{z}_{i}\right)/\tau\right)}{\sum_{x\in\Lambda^{-}}\exp\left(\sin\left(\tilde{z}_{i},g(f(x))\right)/\tau\right)}\end{equation}
where $\tau$ denotes the temperature parameter, and \(\sin(\cdot)\) is the similarity function, which represents the similarity measure between two feature vectors. $\Lambda^{-}$ denotes \(2(N-1)\) negative samples in addition to the positive samples.

\begin{figure*}
\begin{center}
  \includegraphics[width=\linewidth]{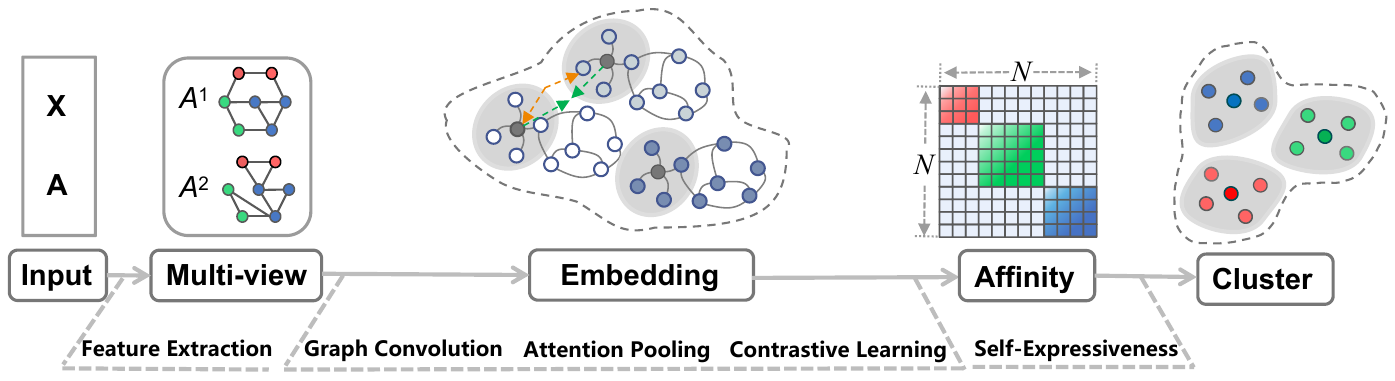}
\end{center}

\caption{Motivation of the MLGSC. First, the spectral-spatial feature views ($A^{1}$) and texture feature views ($A^{2}$) are constructed based on the input HSI data, and the data augmentation is performed on the dual views to obtain the node representations of the dual views using GCN and the global representations of the views through attention pooling. Then, based on the node representations and global representation of the two feature views, multi-level graph contrastive Learning is performed to obtain the joint local-global graph representation. Finally, spectral clustering is applied to the affinity matrix.}

\label{fig:main}
\end{figure*}

%
%

\section{Method}
\label{se:method}

As shown in Figure \ref{fig:main}, our proposed framework consists of five key components: a multi-view graph construction module, a multi-level graph contrastive learning module, an attention pooling module, a graph convolution module, and a self-expression module. First, we introduce the multi-view construction module and then address the specific implementations of the other modules.

\subsection{Multi-view Graph Construction Module}

The primary objective of the multi-view construction module is to extract texture features and spectral features, and then construct multi-views based on these two features. First, since high-dimensional HSIs contain a large number of redundant bands, we use principal component analysis (PCA) \cite{PCA} to downscale the HSI bands. Then, we erode the image using morphological algorithms (open/closed operations) while preserving texture features. In another branch, we extract the pixel and its neighboring pixel information using a sliding window segmentation algorithm to obtain the spatial information of the image. Since then, we can obtain the multi-view features \(\{X'_k\}_{k=1}^2\), where \(X'_k \in \mathbb{R}^{w \times w \times d_k \times n}\), and the multi-view structure can be defined as \(G_k=(V_k, E_k, X_k)\), where \(V_k\) and \(E_k\) represent the corresponding node sets and edge sets, respectively.

Considering that GCN can only work directly on graph structures, we capture the nearest k neighbors of each data and transform the processed data into topological graph structures. In each view,  an adjacency matrix is constructed by computing the Euclidean distance between different samples denoted as \(A_k\). The elements in the adjacency matrix can be described as:
\begin{equation}
A_{i j}^p=\left\{\begin{array}{l}
1, \mathrm{x}_j^p \in \mathcal{N}_k\left(\mathrm{x}_i^p\right) \\
0, \text { otherwise }
\end{array}\right.,
\end{equation}
where $x_{k_i}$ and $x_{k_j}$ are the columns of $A_{k_{ij}}$. The columns of $\mathcal{N}_k(x_{k_i})$ denote the $k$-th view that includes $x_i$ in the neighborhood set sum. $A_{k_{ij}}$ contains the values of $x_{k_i}$ and $x_{k_j}$, indicating whether they are similar or not. Considering that multiple positive and negative samples are required for subsequent multi-level contrastive learning, we perform data augmentation on the two feature views by randomly deleting edges. Specifically, we define the enhanced adjacency matrix as $\widetilde{A}_k$, which is obtained by the following way:

\begin{equation}\tilde{A}_{k_{ij}}=A_{k_{ij}}\times(1-\delta_{ij}),
\end{equation}
where $\delta_{(i,j)}$ is a probability parameter. By this method, we can obtain two spectral views and two texture views after enhancement, updating the multi-view set as $\{X_k\}_{k=1}^4$.

\subsection{Graph Convolution Module}

In HSI clustering tasks, topological relationships among learned objects are often overlooked. Traditional convolutional neural networks can only capture the direct near-neighbor relationships while ignoring the deeper dependencies between graph nodes. Unlike convolutional neural networks, graph convolutional neural networks can make full use of the feature information of each neighboring node and the dependencies between nodes through graph convolution operations. In mathematics, graph convolution is usually defined as:
\begin{equation}
\begin{aligned}
H &= f(X^T, \widetilde{A}; W) \\
&= \sigma\left(\widetilde{D}^{-1/2} \widetilde{A} \widetilde{D}^{-1/2} X^T W\right)
\end{aligned}
\end{equation} 
where $\widetilde{A} = I_N + A$ is a self-cycling adjacency matrix, and $I_N$ is the identity matrix. $\widetilde{D}$ and $W$ are the node degree matrix and weight matrix of each view, respectively. $\sigma$ is the ReLU nonlinear activation function, and $H$ denotes the generative graph embedding. In this model, we have the node feature matrices of the data-enhanced feature views $X$ and the adjacency matrix $A$ as inputs to the GCN. There are two encoders in this model, and the two views from the same feature after data enhancement share the same GCN encoder, i.e., they share a common weight matrix $W$.

\subsection{Attention Pooling to Get Global Graph Representation Module}

One of the key steps in the preparatory phase of multi-level contrastive learning is to distill the global graph representation from each view. This study employs an attention pooling layer, which computes the overall graph representation by weighting the node representations of each view. This process employs an attention mechanism to weight the node representations, emphasizing each node's contribution to the global graph representation.

Specifically, for the set of node features of the $k$-th view $[z_{k,1}, \ldots, z_{k,N}]$, we first transform the features through a linear transformation layer to obtain an attention score for each node. Subsequently, these scores are computed via the softmax normalization process, which are transformed into the node's attention weights.

Mathematically, this process can be expressed as follows: the attentional weights of the nodes $\alpha^P$ by multiplying the set of node features with the weight matrix $M$, applying the $\tanh$ function and softmax function is obtained after normalization. Then, we apply these attention weights to the corresponding node features and compute the global graph representation by weighted summation $S_k$. The specific mathematical expressions are as follows:
\begin{equation}\alpha^P=softmax\left(tanh([z_{k,1},...,z_{k,N}]\cdot M)\right)\end{equation}

\begin{equation}S_k=\sum_{i=1}^N\alpha_i^P\cdot z_{k,i}\end{equation}

The attention pooling process ensures that the global graph representation $S_k$ not only captures the overall nature of the graph structure but also reflects the relative importance of each node in the graph, providing a detailed and informative representation for subsequent graph-level contrastive learning.

\subsection{Multi-level contrastive learning modules}

How to effectively and comprehensively utilize multi-view information is a major challenge for multi-view clustering. That is, we need to learn consistent information from different views to facilitate more robust clustering. With the above GCN, we have extracted the multi-view depth features, and for ease of representation, we denote the feature of a node as $Z$ and the graph-level representation of a view as $S$.

In recent years, contrastive learning has shown excellent performance in unsupervised classification tasks because it can maintain consistency across views by constraining the distribution of similarities and dissimilarities. The idea of contrastive learning is mainly to construct positive and negative samples, minimize the distance between the positive samples, and at the same time, maximize the distance between the positive and negative samples.

However, some recent contrastive learning methods only focus on node-level contrastive learning, disregarding the utilization of global graph information. Furthermore, certain approaches solely concentrate on intra-view contrastive learning, overlooking the potential contributions of inter-view graph contrastive learning effects. Therefore, in this model, we fully consider node-level contrastive learning and graph-level contrastive learning, and at the same time, we make full use of the variability between views from different features.

For any node $i$ on the graph, we treat the node representation in one of the views, denoted as $z_i$ as an anchor point, and the corresponding node representation in the other views $z'_i$ is denoted as positive samples. The other nodes in the same view are denoted as $z_j~(i \neq j)$, and the corresponding nodes in the other views are denoted as $z'_j$. The $z_j$ and $z'_j$ are all denoted as negative samples. Thus, the node-level contrastive loss can be denoted as:
\begin{equation} \label{eq:closs}
\ell(z_i,z_i^{\prime})=-\log\frac{e^{\left(sin\left(z_i,z_i^{\prime}\right)/\tau\right)}}{\sum_{k=1}^Ne^{\left(sin\left(z_i, z_i^{\prime}\right)/\tau\right)}+\sum_{k=1}^Ne^{\left(sin\left(z_i, z_k\right)/\tau\right)}},
\end{equation}
where $\tau$ denotes the temperature parameter and it is fixed to 0.05 in this paper, $\sin(\cdot)$ denotes the similarity function. The overall objective of minimization is the average of Eq. (\ref{eq:closs}) for all given positive samples, which has the following expression:
\begin{equation}\mathcal{L}_{Con}=\frac{1}{2N}\sum_{i=1}^{N}[\ell(z_{i},z_{i}^{\prime})+\ell(z_{i}^{\prime},z_{i})].
\end{equation}

Node-level intra-view and inter-view contrastive loss are defined as $\mathcal{L}_{C_{\text{node}}}$ and $\mathcal{L}_{D_{\text{node}}}$ respectively. Similar to the principle of the node-level contrastive loss function, we define the graph-level contrastive loss function as:
\begin{equation}
\begin{aligned}
&\mathcal{L}_{\text{Graph}}\left(s_{i},s_{j}\right)= \\
&-log\frac{\exp\left(\left(s_{i}^T s_{j}\right)/\tau\right)}{\sum_{\tilde{s}_i}exp\left(\left(s_{i}^T\tilde{s}_{\boldsymbol{i}}\right)/\tau\right)+\sum_{\tilde{s}_j} \exp \left(\left(\boldsymbol{s}_{\boldsymbol{i}}^T\tilde{\boldsymbol{s}}_{\boldsymbol{j}}\right)/\tau\right)}.
\end{aligned}
\end{equation}

\begin{figure}
\begin{center}
  \includegraphics[width=\linewidth]{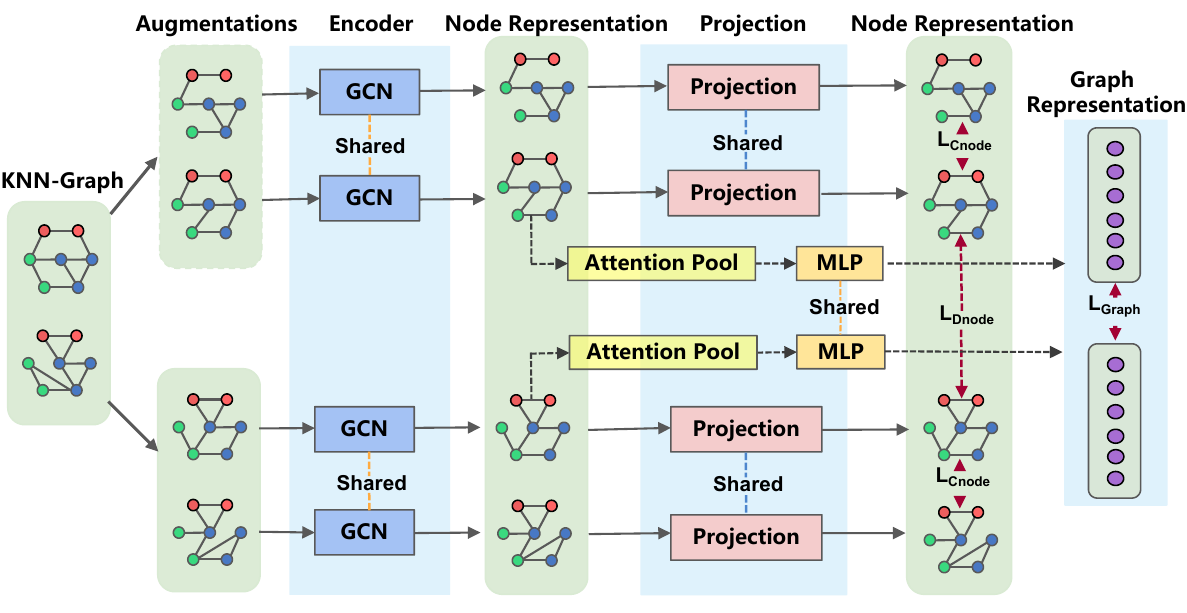}
\end{center}
\caption{Multi-level Contrastive Learning Framework.}
\label{fig:2}
\end{figure}

\subsection{Weighted Feature Fusion Module}

After processing through the GCN and the multi-level contrastive learning module, we obtain multi-view representations. Effectively fusing these representations is essential for maximizing their potential. Our objective is to derive a discriminative fusion feature representation, denoted as $X_s$, which captures the comprehensive essence of all views. To achieve this, we employ a strategy of weighted feature fusion, ensuring optimal integration of these diverse features.

Given the set of computed feature representations $I$, we first construct the weight mapping matrix $m_k$. It is expected that feature representations extracted from different features have different fusion weights at the graph node level. Therefore, we introduce a modal-level and node-level softmax function from $I$ generating the weight mapping matrix. In this context, for the $i$-th node of the graph, $z_k'$ and $m_k$ are denoted as $z_k^i$ and $m_k^i$; the weight mapping matrix $m_k$ can be defined as:
\begin{equation}m_k^i=\frac{e^{z_k^i}}{\sum_{j\in K}e^{z_j^i}}\end{equation}

By multiplying the input feature maps element by element $X_k$ with the corresponding weight matrix $m_k$ and then summing over all the features, we can obtain a fused feature map:
\begin{equation}F_s=\sum_{k\in K}X_k\cdot m_k\end{equation}

Since the sum of $m_1^i, \ldots, m_{|K|}^i$ sums to 1, the fusion feature representation $F_s$ of the range is stabilized to improve stability against variable modal inputs.

\subsection{Self-Expression Module}
Inspired by the graph convolutional subspace clustering method \cite{12cai}, we improve the traditional self-expression coefficient matrix to obtain more robust information, and reconstruct the feature expression as follows:
\begin{equation}X=X\bar{A}C,\quad \text { s.t. diag }(C)=0\end{equation}
where $C$ denotes the matrix of self-expression coefficients for each view, $\bar{A}$ denotes the normalized adjacency matrix, and $X$ is the initial value of the feature after attention feature fusion $X_s$. The final graph self-expression can be defined as:
\begin{equation}\underset{C}{\arg\min}\frac{1}{2}\|\mathrm{X}\overline{\mathrm{A}}C-\mathrm{X}\|_{\mathrm{F}}^2+\frac{\lambda}{2}\|C\|_{\mathrm{F}}^2 \quad \text { s.t. diag }(C)=0
\end{equation}
where $\|\cdot\|_F$ is the Frobenius norm, and $\lambda$ is the equilibrium coefficient. Since the self-expression dictionary matrix $Z = X\bar{A}$ contains global structural information, a clearer dictionary can be obtained to generate robust affinity matrices. Thus, the graph convolution self-expression is often utilized to reconstruct the original data using the self-expression dictionary matrix $Z$. The self-expression loss function is defined as follows:
\begin{equation}\begin{aligned}
\mathcal{L}_{S E}& =\frac{1}{2}(\|ZC-X\|_{F}+\lambda\|C\|_{F})  \\
&=\frac12tr\left[(ZC-X)^T(ZC-X)+\lambda C^TC\right]
\end{aligned}\end{equation}

We base the multi-view self-expression coefficient matrix $C$ to construct the affinity matrix $W$ for the subsequent implementation of spectral clustering with the following affinity matrix expression:
\begin{equation}W=\frac12(|C|+|C|^\mathrm{T})\end{equation}

\subsection{Total Loss Parameter Optimization Module}

Considering that there are multiple loss functions in this model, how to weigh the weights between the losses will have a direct impact on the final clustering results of the model.

According to the content of each module described earlier, we can get two node-level contrastive losses and one graph-level contrastive loss, respectively, multiply them by the corresponding weights, sum up, we can get the total loss $\mathcal{L}$ for:
\begin{equation}
\begin{aligned}
\mathcal{L}= 
&\frac{1}{2\sigma_{1}^{2}}\mathcal{L}_{Cnode}+
\frac{1}{2\sigma_{2}^{2}}\mathcal{L}_{Dnode}+ 
\frac{1}{2\sigma_{3}^{2}}\mathcal{L}_{Graph}+ \\
&\frac{1} {2\sigma_{4}^{2}}\mathcal{L}_{SE}+
\log(\prod_{i=1}^{4}\sigma_i),
\end{aligned}
\end{equation}
where $\{\sigma_i, i=1,2,3,4\}$ denotes the trainable variance of each task. The larger the $\sigma_i$, the greater uncertainty of the task, and the corresponding task weights decrease proportionally. Our model co-optimizes the model parameters and these variances in an end-to-end manner using an adaptive moment estimation optimizer. To improve the stability of the model parameter range, we include the regularization term $\log\left(\prod_{i=1}^{4} \sigma_i\right)$.

\begin{table*}[!t]
\caption{Clustering Results (OA/NMI/KAPPA) of Various Methods on Four Benchmark Datasets. The Best and Suboptimal Results are Shown in Bold and Underlined, Respectively.}
\label{tb_main}
\renewcommand\arraystretch{1.3}
\begin{tabular}{ccccccccccccc} \hline
\textbf{Dateset} & \textbf{Metric} & \textbf{k-means} & \textbf{SSC} & \textit{\textbf{$l_2$-SSC}} & \textbf{RMMF} & \textbf{NMFAML} & \textbf{EGCSC} & \textbf{GCOT} & \textbf{HGCSC} & \textbf{GR-RSCNet} & \textbf{DMISC} & \textbf{MLGSC} \\ \hline
\multirow{3}{*}{\textbf{InP.}} & OA & 0.5962 & 0.6562 & 0.6641 & 0.7125 & 0.8512 & 0.8486 & 0.8451 & 0.9246 & 0.9315 & {\ul 0.932} & \textbf{0.9775} \\
 & NMI & 0.4478 & 0.6106 & 0.5381 & 0.4984 & 0.7261 & 0.6441 & 0.7654 & 0.8921 & 0.8347 & {\ul 0.839} & \textbf{0.9288} \\
 & Kappa & 0.4412 & 0.5527 & 0.5269 & 0.5607 & 0.7863 & 0.6422 & 0.6385 & 0.8312 & 0.9013 & {\ul 0.902} & \textbf{0.9677} \\ \hline
\multirow{3}{*}{\textbf{PaU.}} & OA & 0.6025 & 0.6542 & 0.6912 & 0.7712 & 0.8964 & 0.8421 & 0.9103 & 0.9531 & {\ul 0.9775} & 0.961 & \textbf{0.9996} \\
 & NMI & 0.5306 & 0.6408 & 0.6257 & 0.7384 & 0.9231 & 0.8405 & 0.8804 & 0.9384 & {\ul 0.9685} & 0.958 & \textbf{0.9997} \\
 & Kappa & 0.6791 & 0.5734 & 0.6883 & 0.6805 & 0.8521 & 0.7968 & 0.9236 & 0.9441 & {\ul 0.971} & 0.948 & \textbf{0.9984} \\ \hline
\multirow{3}{*}{\textbf{Hou.}} & OA & 0.3928 & 0.4582 & 0.5110 & 0.7310 & 0.6347 & 0.6236 & - & 0.7314 & 0.8389 & {\ul 0.893} & \textbf{0.9228} \\
 & NMI & 0.4123 & 0.4963 & 0.5613 & 0.7789 & 0.7965 & 0.7745 & - & 0.7014 & 0.8729 & {\ul 0.877} & \textbf{0.9259} \\
 & Kappa & 0.3287 & 0.4015 & 0.4813 & 0.6213 & 0.5921 & 0.5931 & - & 0.8307 & 0.8231 & {\ul 0.885} & \textbf{0.9142} \\ \hline
\multirow{3}{*}{\textbf{XuZ.}} & OA & 0.5589 & 0.6631 & 0.6836 & 0.6145 & 0.8145 & 0.7925 & - & 0.9154 & {\ul 0.9193} & - & \textbf{0.9573} \\
 & NMI & 0.4123 & 0.5352 & 0.5967 & 0.6251 & 0.7919 & 0.6958 & - & {\ul 0.8863} & 0.8741 & - & \textbf{0.9086} \\
 & Kappa & 0.4172 & 0.4762 & 0.6593 & 0.4362 & 0.7459 & 0.6482 & - & 0.8476 & {\ul 0.8642} & - & \textbf{0.9423} \\ \hline
\end{tabular}
\end{table*}

\section{Experiments}
\label{se:experiments}
In this section, we evaluate the clustering performance of MLGSC on four benchmark datasets and compare it with several state-of-the-art clustering models. 

\subsection{Experiment Setup}
\textbf{Datasets.} We conduct experiments on four commonly used HSI datasets, including the Indian Pines dataset, the Pavia University dataset, the Houston-2013 dataset, and the Xu Zhou dataset, to ensure that our models are fairly evaluated across scenarios. The Indian Pines dataset is acquired by the AVIRIS sensor in Indian Pines in northwestern Indiana, containing 16 land cover types with a spatial resolution of 20 m. The experimentally selected sub-scene is located at [30-115, 24-94]. The Pavia University dataset is acquired by the German ROSIS sensor, containing 9 land cover types with a spatial resolution of 1.3 m. The experimentally select sub-scenes are located at [50-350, 100-200]. Houston-2013 dataset is acquired by ITRES CASI-1500 sensor and contains 12 land cover types. The size of the dataset is 349*1905, and the experimentally selected sub-scenes are located at [0-349,0-680]. The Xu Zhou dataset is acquired by the high-resolution 0.73 m airborne HySpex imaging spectrometer and contains 5 land cover types. The dataset contains 436 bands with a spectral range of 415-2508 nm, and the experimentally selected sub-scenes are located at [0-100,0-260].

\textbf{Baseline Methods.} We compare MLGSC with ten representative clustering methods, including traditional clustering algorithms as well as several excellent clustering algorithms developed in recent years: K-means \cite{33kanungo2002efficient}, SSC \cite{8zhang2016spectral}, SSC with the introduction of $l_2$-paradigm number ($l_2$-SSC) \cite{35zhai2016new}, Robust Matrix Decomposition of Flows (RMMF) based method \cite{36zhang2019hyperspectral}, Graph Convolutional Subspace Clustering Algorithm (EGCSC)\cite{12cai}, GCOT, HGCSC, Graph Regularized Residual Subspace Clustering Network (GR-RSCNet) \cite{17cai2021graph}, and Deep Mutual Information Subspace Clustering Network (DMISC) \cite{24li2022deep}. We follow the settings suggested in the corresponding articles.

\textbf{Evaluation Metrics.} In order to quantitatively evaluate the performance of MLGSC and comparison methods, we use three commonly used evaluation metrics, overall accuracy (OA), normalized mutual information (NMI), and Kappa coefficient (Kappa), to evaluate the models proposed in this paper. These metrics range from [0,1] (OA and NMI) or [-1,1] (Kappa), with higher scores indicating more accurate clustering results.

\subsection{Quantitative Results}

Table \ref{tb_main} shows the clustering performance of MLGSC and other models for these four datasets. Our model achieves the best clustering performance for most of the three evaluation metrics, OA, NMI, and Kappa, significantly outperforming the other methods. Specifically, for the Indian Pines, Houston, and Xu Zhou datasets, the OA of the method improves by 4.5-6.5\% compared to the sub-optimal methods. In addition, we observe the following trends:

(1) Obtaining a global graph representation through attention pooling can lead to a significant improvement in the global perception of features. This enhancement is primarily due to the attention mechanism's ability to weigh different parts of the graph according to their relevance, allowing the model to focus on more informative regions while disregarding less important ones. Attention pooling dynamically aggregates node features by considering their contextual importance, leading to a more comprehensive and nuanced global graph representation. 

(2) Combined consideration of multi-view node-level contrastive learning and graph-level contrastive learning can effectively improve the clustering accuracy. Node-level contrastive learning can utilize the structural information within each view, while graph-level contrastive learning can fully consider the variability among views from different features. Our model fully utilizes the structural information within each view and the complementary information among multi-views, and achieves 97.75\%, 99.96\%, 92.28\%, and 95.73\% accuracy on Indian Pines, Pavia University, Houston-2013, and Xu Zhou datasets, respectively.

(3) Our model outperforms other methods in terms of experimental accuracy on all four datasets. Specifically, our model improves 12.52\%, 14.96\%, 26.14\%, and 15.66\% on all datasets, respectively. This suggests that the introduction of an attention pooling module and the simultaneous use of view internal structure information and complementary information among global multi-views can effectively improve the clustering accuracy, and our model provides a novel solution for HSI clustering.

\begin{figure}
\begin{center}
  \includegraphics[width=\linewidth]{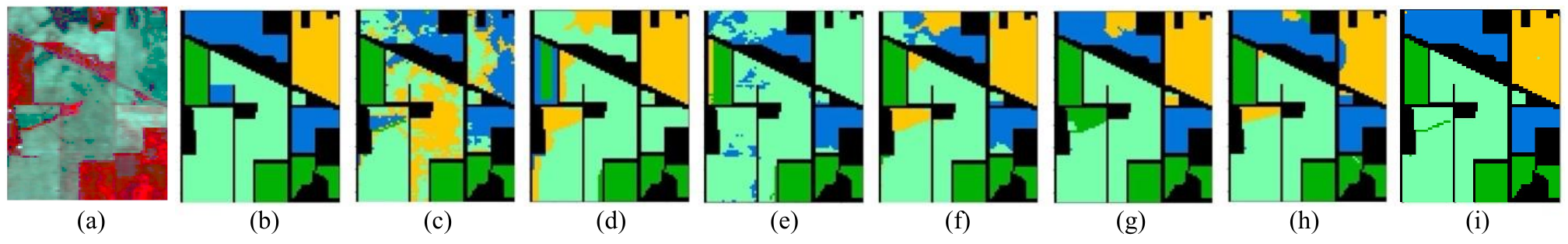}
\end{center}
\caption{Clustering visualization on Indian Pines dataset: (a) original image, (b) ground truth, (c) k-means, (d) SSC, (e) 12-SSC, (f) EGCSC, (g) HGCSC, (h) GR-RSCNet, (i) our proposed MLGSC.}
\label{fig:3}
\end{figure}

\begin{figure}
\begin{center}
  \includegraphics[width=\linewidth]{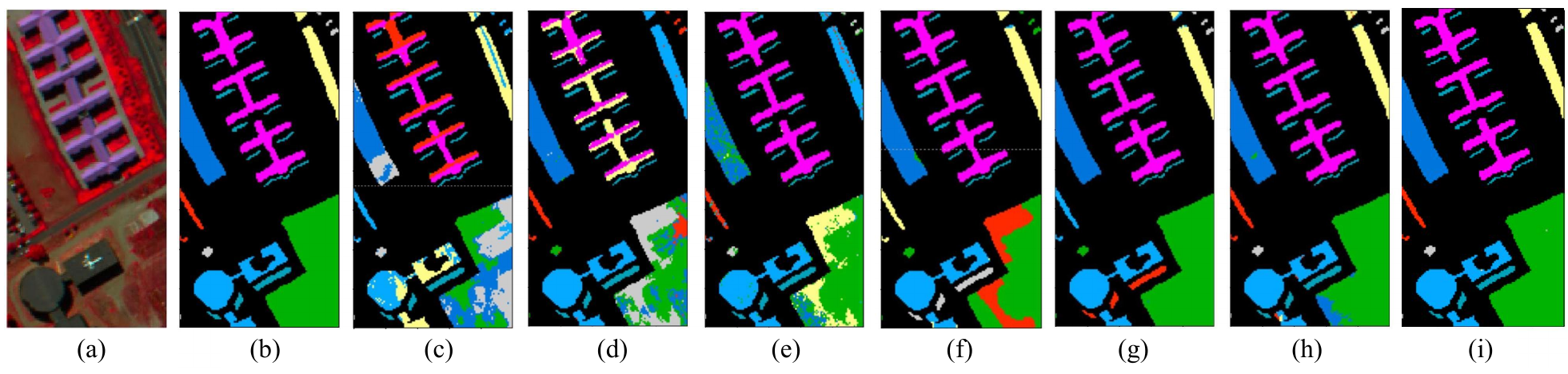}
\end{center}
\caption{ Clustering visualization of Pavia University dataset: (a) original image, (b) ground truth, (c) k-means, (d) SSC, (e) 12-SSC, (f) EGCSC, (g) HGCSC, (h) GR-RSCNet, (i) our proposed MLGSC.}
\label{fig:4}
\end{figure}

\begin{figure}
\begin{center}
  \includegraphics[width=\linewidth]{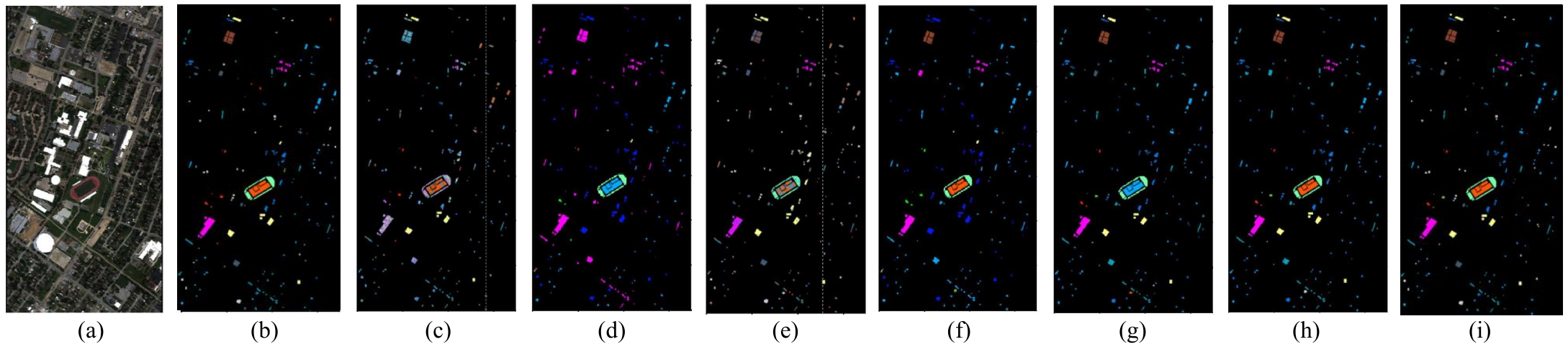}
\end{center}
\caption{ Clustering visualization of Houston2013 dataset: (a) original image, (b) ground truth, (c) k-means, (d) SSC, (e) 12-SSC, (f) EGCSC, (g) HGCSC, (h) GR-RSCNet, (i) our proposed MLGSC.}
\label{fig:5}
\end{figure}

\begin{figure}
\begin{center}
  \includegraphics[width=\linewidth]{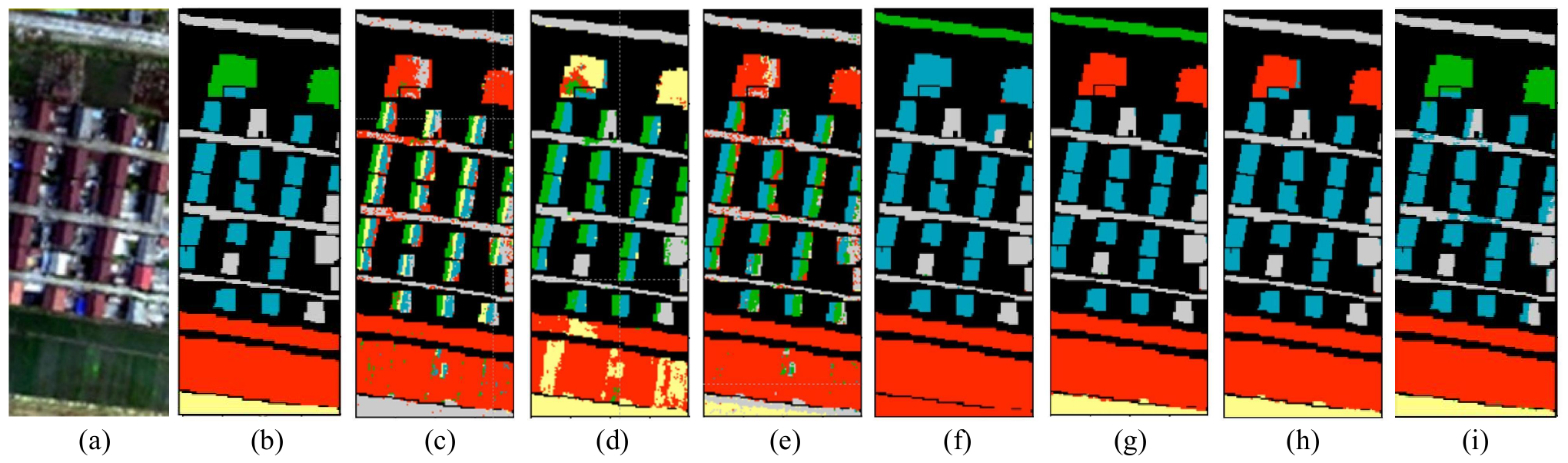}
\end{center}
\caption{ Clustering visualization of Xuzhou dataset: (a) original image, (b) ground truth, (c) k-means, (d) SSC, (e) 12-SSC, (f) EGCSC, (g) HGCSC, (h) GR-RSCNet, (i) our proposed MLGSC.}
\label{fig:6}
\end{figure}

\subsection{Comparison With State-of-the-Art Methods}

To make a fair comparison between the different methods, we reproduce the baseline under the same data preprocessing as MLGSC. We conclude that MLGSC significantly outperforms all other baseline methods on all datasets. For example, MLGSC improved over baseline on the Indian Pines (OA = 97.75\%), Pavia University (OA = 99.96\%), Houston2013 (OA = 92.28\%), and Xu Zhou (OA = 95.73\%) datasets by at least 4.75, 14.56, and 5.68\%. This demonstrates the effectiveness of the MLGSC method.

Figure \ref{fig:3}-\ref{fig:6} show the visualization results of different clustering methods on Indian Pines, Pavia University, Houston2013, and Xuzhou datasets, respectively. Each image of (a) represents the original image of the HSI, and (b) shows the ground truth of the corresponding dataset after removing the irrelevant background. SSC can extract low-dimensional information from high-dimensional data structures but does not consider spatial constraints. In addition, traditional subspace clustering methods have difficulty in accurately modeling the structure of HSI, which makes them inferior to deep learning-based methods. Our model achieves the best visualization results on all datasets. MLGSC has smaller pretzel noise in clustered images due to the adaptive aggregation of texture information. Also, since the GCN model can aggregate neighboring node information, better homogeneity is maintained within the same class of regions and the image boundaries remain relatively intact.

\begin{figure}
\begin{center}
  \includegraphics[width=\linewidth]{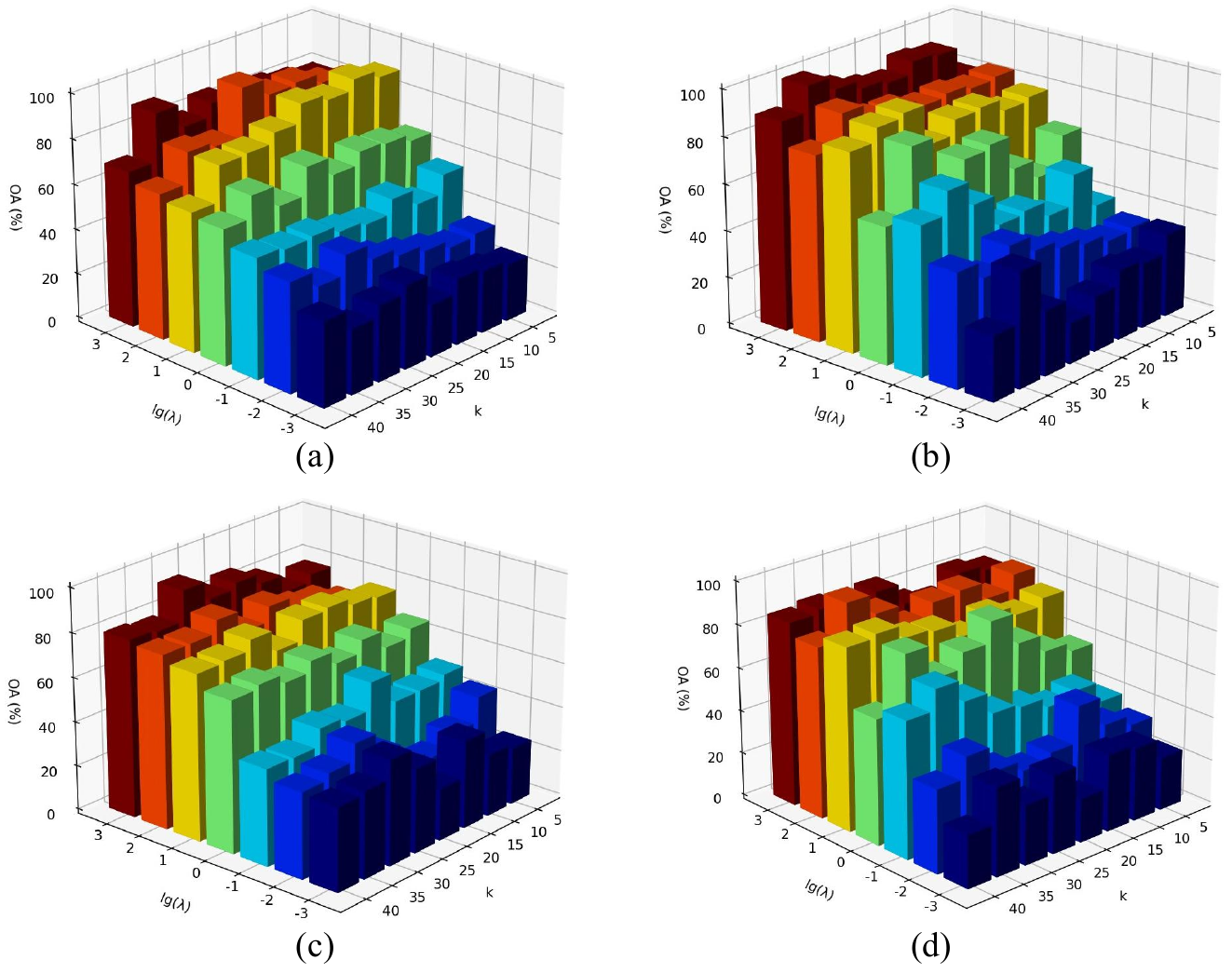}
\end{center}
\caption{On four datasets $\lambda$ and $k$: (a) Indian Pines, (b) Pavia University, (c) Houston2013, (d) Xu Zhou.}
\label{fig:7}
\end{figure}

\begin{figure}
\begin{center}
  \includegraphics[width=8cm]{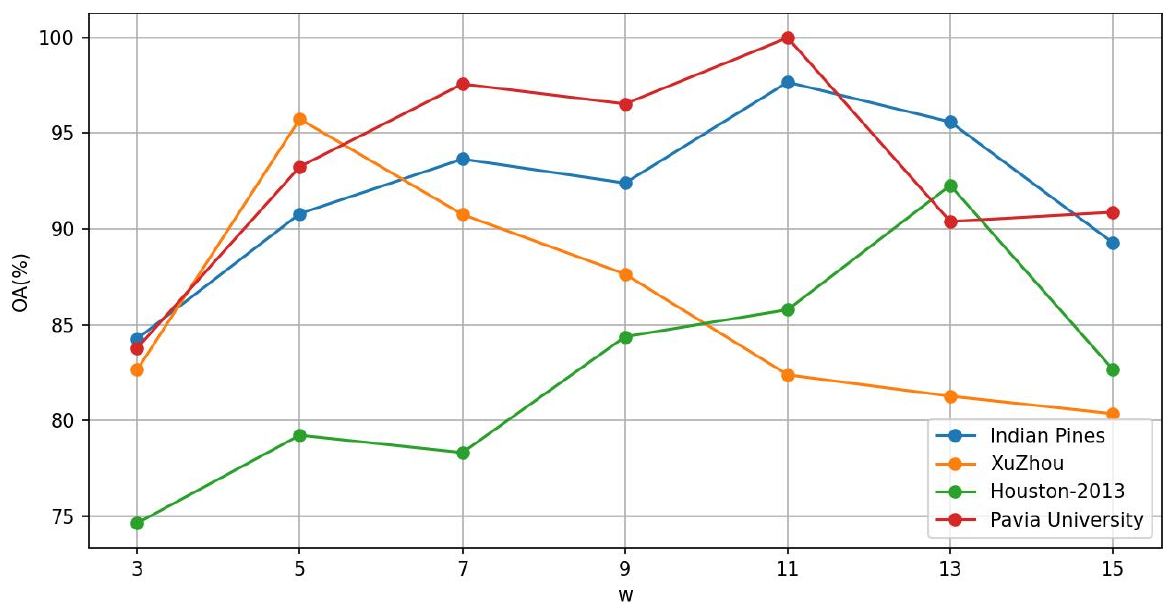}
\end{center}
\caption{The performance comparison of the four datasets with different patch window sizes $w$.}
\label{fig:8}
\end{figure}

\begin{figure}
\begin{center}
  \includegraphics[width=8cm]{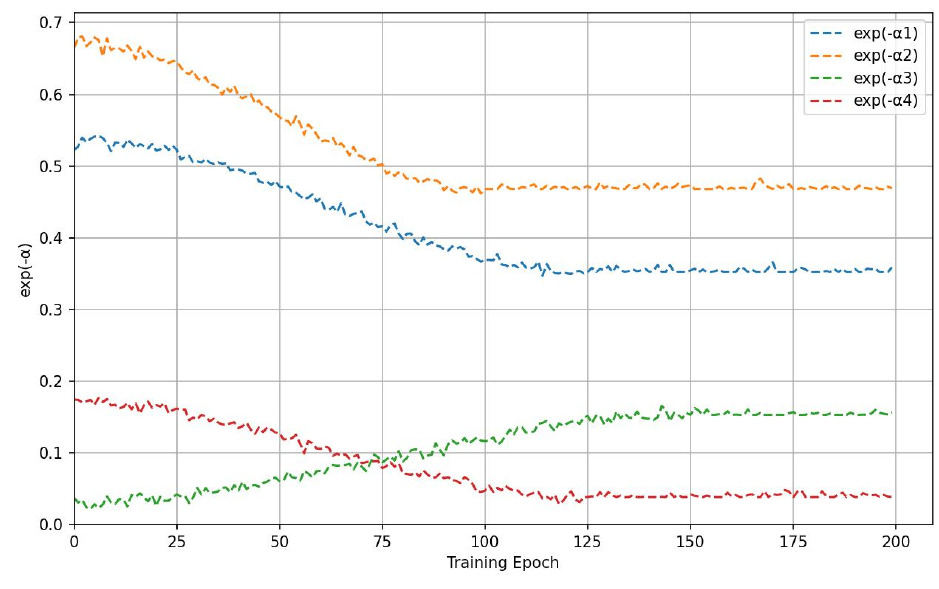}
\end{center}
\caption{Evolution of loss weights over training epochs on the Indian Pines dataset, where the loss weight is indicated as $\exp(-\alpha_i)$ with $\alpha_i = \log(\sigma_i)$. (a) Weight of $L_{C_{\text{node}}}$. (b) Weight of $L_{D_{\text{node}}}$. (c) Weight of $L_{\text{Graph}}$. (d) Weight of $L_{\text{SE}}$.
}
\label{fig:9}
\end{figure}

\begin{table*}[] 
\caption{Ablation Studies (OA) for Different Views, Contrastive Learning, and Attention Pooling Module for MLGSC.}
\label{tb:ablation}
\renewcommand\arraystretch{1.3}
\resizebox{\textwidth}{!}{
\begin{tabular}{cccccccccc} \hline
\multirow{2}{*}{No.} & \multirow{2}{*}{Attention Pooling} & Contrastive Learning & Contrastive Learning & \multirow{2}{*}{Texture Features} & \multirow{2}{*}{Space-spectral   Features} & \multirow{2}{*}{InP} & \multirow{2}{*}{PaU} & \multirow{2}{*}{Hou} & \multirow{2}{*}{XuZ} \\ 
 &  & (Node-level) & (Graph-level) &  &  &  &  &  &  \\ \hline
1 & × & $\checkmark$ & $\checkmark$ & $\checkmark$ & $\checkmark$ & 0.9567 & 0.9431 & 0.8725 & 0.9264 \\
2 & $\checkmark$ & × & $\checkmark$ & $\checkmark$ & $\checkmark$ & 0.8815 & 0.8747 & 0.8066 & 0.8421 \\
3 & $\checkmark$ & $\checkmark$ & × & $\checkmark$ & $\checkmark$ & 0.9236 & 0.9151 & 0.8517 & 0.9133 \\
4 & $\checkmark$ & $\checkmark$ & $\checkmark$ & × & $\checkmark$ & 0.8319 & 0.7964 & 0.7537 & 0.8233 \\
5 & $\checkmark$ & $\checkmark$ & $\checkmark$ & $\checkmark$ & × & 0.9281 & 0.9188 & 0.7942 & 0.8029 \\
6 & $\checkmark$ & $\checkmark$ & $\checkmark$ & $\checkmark$ & $\checkmark$ & 0.9775 & 0.9996 & 0.9228 & 0.9573 \\ \hline
\end{tabular}}
\end{table*}

\subsection{Parameter Analysis}
In this study, we investigated the effects of the three most important hyperparameters on the MLGSC framework. 

\textbf{Regularization coefficients $\lambda$ sensitivity.} We choose $\lambda$ from [0.01, 0.1, 1, 10, 100, 1000], and through a series of experiments, we find that smaller values of $\lambda$ resulted in lower OA, NMI, and Kappa values. This indicates that the smaller the value of $\lambda$, the less the model is penalized and the easier it is to overfit the training data. This means that the model is more inclined to over-rely on the details and noise of the training data rather than a more generalized clustering structure. As a result, this situation may lead to poorer generalization performance of the model. On the Pavia University dataset, MLGSC still obtains near-optimal OA, NMI, and Kappa metrics when varying the value of $\lambda$, which suggests that our method is robust to regularization coefficients.

\textbf{Sensitivity of nearest neighbor $k$.} Smaller values of $k$ may lead to a model that is sensitive to noise, while larger values of $k$ may lead to a model that is too smooth and ignores the local clustering structure, so it is crucial to choose the appropriate $k$ value. We chose the effect of $k$ in the range of values between 20 and 40 while keeping other parameters constant. Through a series of experiments we find that larger values of $k$ resulted in lower OA, NMI and Kappa, which suggests that larger values of $k$ may cause the model to over-average the overall structure and ignore local clustering features in the data. Considering all kinds of situations, we set the $k$-values of the four datasets to 25, 35, 30, and 35, respectively.

\textbf{Sensitivity to patch window size $w$.} $w$ is the size of the spatial window used to extract spatial information, which has a certain degree of influence on the data for local information capture and model performance tuning. In the experimental process, we set the interval of the value of $w$ to [5,13] and gradually increase 2. The experimental results are shown in Figure \ref{fig:8}. We find that if the window size is chosen to be too small, it may contain more redundant spatial information, which leads to a loss of accuracy. This indicates that the model fails to fully consider the overall structure of the data during the clustering process and focuses too much on the detailed information of local neighborhoods. This may lead to instability and inaccuracy of the clustering results, making the model susceptible to noise and overfitting. MLGSC works best on these four datasets when the parameter $w$ is 11, 11, 13, and 5, respectively.

\textbf{Convergence analysis.} In Figure \ref{fig:9}, we observe the convergence of the MLGSC model by visualizing the clustering metric on the loss values of 200 training cycles. In the initialization phase, we use a matrix with all 1s to initialize the self-expressiveness layer, at which time the model does not represent the data accurately enough, so the initial clustering of the model is poor. However, as the training period increases, we can observe that the clustering metrics increase significantly and then stabilize, while the loss values tend to converge. Eventually, the four metrics converge at 0.352, 0.468, 0.153, and 0.038, respectively.  This indicates that as training proceeds, the model gradually adjusts the parameters and learns a more effective representation, thus improving the clustering performance.

\subsection{Ablation Study}
We performed various ablation experiments on each module in MLGSC to consider the contribution of each module to the structural soundness of the model. The quantitative results of the model on the Indian Pines, Pavia University, Houston-2013 and Xu Zhou datasets are given in Table \ref{tb:ablation}, and the results are averaged over 10 experiments.

\textbf{Ablation experiments on attention pooling module.} Table \ref{tb:ablation} compares the clustering accuracies with and without the attention pooling module. We observe that the model with the attentional pooling module outperforms the simplified model without the attentional pooling module on all datasets and achieves higher OA, NMI, and Kappa. The addition of the attention pooling module improves the models for the four datasets by 2.9\%, 3.76\%, 4.51\%, and 6.09\%, respectively, which demonstrates that the attention pooling module significantly improves the clustering performance.

\textbf{Ablation experiments on multi-level contrastive learning.} We consider three contrastive patterns, including contrastive between the same feature view nodes, contrastive between different feature view nodes, and graph-level contrastive between different feature views, which are important components of the proposed contrastive patterns. Table \ref{tb:ablation} compares the clustering effects of only the comparison between the same feature view nodes, only the graph-level comparison between different feature views, and the multi-level comparison mode. The comparison results show that the multi-level comparison mode proposed in this paper outperforms its component comparison mode, further indicating that better performance can be achieved synergistically by considering both local and global comparison modes together.

\textbf{Ablation experiments on multi-views.} We have utilized a combination of texture features and spectral-spatial features of HSI data to confirm the superiority of constructing texture feature views and spectral-spatial feature views at the same time. Table \ref{tb:ablation} compares the experimental results in three cases, i.e., constructing only the texture feature views, constructing only spectral-spatial feature views, and constructing both the texture view and spectral-spatial feature views at the same time. We observe that the experimental performance of simultaneously constructing texture views and spectral-spatial feature views is significantly better than the other two cases on all datasets, and this result strongly demonstrates the importance of considering both texture features and spectral-spatial features.

\section{Conclusion}
\label{se:conclusion}
In this study, we introduce the MLGSC model, an advanced multi-level graph contrastive learning subspace clustering network for HSI clustering. The MLGSC model deeply mines the potential features of the HSI data through the multi-view building module and cleverly applies attention pooling to obtain the global graph representation, an innovative means that significantly enhances the global perception of features. Meanwhile, we propose a multi-level graph contrastive learning strategy that integrates node-level contrastive learning and graph-level contrastive learning, which further promotes the effective fusion of information between different feature views and enhances the consistency and complementarity of model learning. After testing on several standard HSI datasets, MLGSC achieves a new level of clustering accuracy, proving the effectiveness of the proposed method. In the future, we plan to optimize the model for larger datasets to expand its potential and coverage in practical applications.

\section{Acknowledgment}
This study was jointly supported by the Natural Science Foundation of China under Grants 42071430 and U21A2013, the Opening Fund of Key Laboratory of Geological Survey and Evaluation of Ministry of Education under Grant GLAB2022ZR02 and Grant GLAB2020ZR14, as well as in part by the College Students' Innovative Entrepreneurial Training Plan Program (202310491003) and the Fundamental Research Funds for National University, China University of Geosciences (Wuhan) (No.CUG DCJJ202227).

Computation of this study was performed by the High-performance GPU Server (TX321203) Computing Centre of the National Education Field Equipment Renewal and Renovation Loan Financial Subsidy Project of China University of Geosciences, Wuhan.


\bibliographystyle{IEEEtran}
\bibliography{IEEEfull,refs}

\begin{thebibliography}{10}
\providecommand{\url}[1]{#1}
\csname url@samestyle\endcsname
\providecommand{\newblock}{\relax}
\providecommand{\bibinfo}[2]{#2}
\providecommand{\BIBentrySTDinterwordspacing}{\spaceskip=0pt\relax}
\providecommand{\BIBentryALTinterwordstretchfactor}{4}
\providecommand{\BIBentryALTinterwordspacing}{\spaceskip=\fontdimen2\font plus
\BIBentryALTinterwordstretchfactor\fontdimen3\font minus \fontdimen4\font\relax}
\providecommand{\BIBforeignlanguage}[2]{{%
\expandafter\ifx\csname l@#1\endcsname\relax
\typeout{** WARNING: IEEEtran.bst: No hyphenation pattern has been}%
\typeout{** loaded for the language `#1'. Using the pattern for}%
\typeout{** the default language instead.}%
\else
\language=\csname l@#1\endcsname
\fi
#2}}
\providecommand{\BIBdecl}{\relax}
\BIBdecl

\bibitem{ResCapsNet}
R.~Guan, Z.~Li, T.~Li, X.~Li, J.~Yang, and W.~Chen, ``Classification of heterogeneous mining areas based on rescapsnet and gaofen-5 imagery,'' \emph{Remote Sensing}, vol.~14, no.~13, p. 3216, 2022.

\bibitem{2yang2022enhanced}
G.~Yang, K.~Huang, W.~Sun, X.~Meng, D.~Mao, and Y.~Ge, ``Enhanced mangrove vegetation index based on hyperspectral images for mapping mangrove,'' \emph{ISPRS Journal of Photogrammetry and Remote Sensing}, vol. 189, pp. 236--254, 2022.

\bibitem{guan4}
R.~Li, Y.~Hu, L.~Li, R.~Guan, R.~Yang, J.~Zhan, W.~Cai, Y.~Wang, H.~Xu, and L.~Li, ``Smwe-gfpnnet: A high-precision and robust method for forest fire smoke detection,'' \emph{Knowledge-Based Systems}, vol. 289, p. 111528, 2024.

\bibitem{4wang2020identification}
Q.~Wang, L.~Sun, Y.~Wang, M.~Zhou, M.~Hu, J.~Chen, Y.~Wen, and Q.~Li, ``Identification of melanoma from hyperspectral pathology image using 3d convolutional networks,'' \emph{IEEE Transactions on Medical Imaging}, vol.~40, no.~1, pp. 218--227, 2020.

\bibitem{Guan2}
Z.~Liu, R.~Guan, J.~Hu, W.~Chen, and X.~Li, ``Remote sensing scene data generation using element geometric transformation and gan-based texture synthesis,'' \emph{Applied Sciences}, vol.~12, no.~8, p. 3972, 2022.

\bibitem{zhai2021hyperspectral}
H.~Zhai, H.~Zhang, P.~Li, and L.~Zhang, ``Hyperspectral image clustering: Current achievements and future lines,'' \emph{IEEE Geoscience and Remote Sensing Magazine}, vol.~9, no.~4, pp. 35--67, 2021.

\bibitem{2024AMGC}
W.~Tu, R.~Guan, S.~Zhou, C.~Ma, X.~Peng, Z.~Cai, Z.~Liu, J.~Cheng, and X.~Liu, ``Attribute-missing graph clustering network,'' in \emph{Proceedings of the Thirty-Eighth AAAI Conference on Artificial Intelligence (AAAI)}, 2024.

\bibitem{5peng}
B.~Peng, Y.~Yao, J.~Lei, L.~Fang, and Q.~Huang, ``Graph-based structural deep spectral-spatial clustering for hyperspectral image,'' \emph{IEEE Transactions on Instrumentation and Measurement}, vol.~72, pp. 1--12, 2023.

\bibitem{wen1}
Z.~Wen, Y.~Ling, Y.~Ren, T.~Wu, J.~Chen, X.~Pu, Z.~Hao, and L.~He, ``Homophily-related: Adaptive hybrid graph filter for multi-view graph clustering,'' \emph{arXiv preprint arXiv:2401.02682}, 2024.

\bibitem{6vidal2011subspace}
R.~Vidal, ``Subspace clustering,'' \emph{IEEE Signal Processing Magazine}, vol.~28, no.~2, pp. 52--68, 2011.

\bibitem{wen5}
C.~Cui, Y.~Ren, J.~Pu, J.~Li, X.~Pu, T.~Wu, Y.~Shi, and L.~He, ``A novel approach for effective multi-view clustering with information-theoretic perspective,'' \emph{Advances in Neural Information Processing Systems}, vol.~36, 2024.

\bibitem{9zhang}
Z.~Zhang, Y.~Cai, W.~Gong, P.~Ghamisi, X.~Liu, and R.~Gloaguen, ``Hypergraph convolutional subspace clustering with multihop aggregation for hyperspectral image,'' \emph{IEEE Journal of Selected Topics in Applied Earth Observations and Remote Sensing}, vol.~15, pp. 676--686, 2022.

\bibitem{wen3}
W.~Yan, Y.~Zhou, Y.~Wang, Q.~Zheng, and J.~Zhu, ``Multi-view semantic consistency based information bottleneck for clustering,'' \emph{Knowledge-Based Systems}, vol. 288, p. 111448, 2024.

\bibitem{wen4}
Y.~Zhou, Q.~Zheng, Y.~Wang, W.~Yan, P.~Shi, and J.~Zhu, ``Mcoco: Multi-level consistency collaborative multi-view clustering,'' \emph{Expert Systems with Applications}, vol. 238, p. 121976, 2024.

\bibitem{10elhamifar2013sparse}
E.~Elhamifar and R.~Vidal, ``Sparse subspace clustering: Algorithm, theory, and applications,'' \emph{IEEE transactions on pattern analysis and machine intelligence}, vol.~35, no.~11, pp. 2765--2781, 2013.

\bibitem{wen2}
C.~Cui, Y.~Ren, J.~Pu, X.~Pu, and L.~He, ``Deep multi-view subspace clustering with anchor graph,'' \emph{arXiv preprint arXiv:2305.06939}, 2023.

\bibitem{35zhai2016new}
H.~Zhai, H.~Zhang, L.~Zhang, P.~Li, and A.~Plaza, ``A new sparse subspace clustering algorithm for hyperspectral remote sensing imagery,'' \emph{IEEE Geoscience and Remote Sensing Letters}, vol.~14, no.~1, pp. 43--47, 2016.

\bibitem{wuda1}
Y.~Chen, Y.~Tang, Z.~Yin, T.~Han, B.~Zou, and H.~Feng, ``Single object tracking in satellite videos: A correlation filter-based dual-flow tracker,'' \emph{IEEE Journal of Selected Topics in Applied Earth Observations and Remote Sensing}, vol.~15, pp. 6687--6698, 2022.

\bibitem{wuda2}
Y.~Chen, Q.~Yuan, Y.~Tang, Y.~Xiao, J.~He, and L.~Zhang, ``Spirit: Spectral awareness interaction network with dynamic template for hyperspectral object tracking,'' \emph{IEEE Transactions on Geoscience and Remote Sensing}, vol.~62, pp. 1--16, 2024.

\bibitem{AMFGCN}
J.~Liu, R.~Guan, Z.~Li, J.~Zhang, Y.~Hu, and X.~Wang, ``Adaptive multi-feature fusion graph convolutional network for hyperspectral image classification,'' \emph{Remote Sensing}, vol.~15, no.~23, p. 5483, 2023.

\bibitem{12cai}
Y.~Cai, Z.~Zhang, Z.~Cai, X.~Liu, X.~Jiang, and Q.~Yan, ``Graph convolutional subspace clustering: A robust subspace clustering framework for hyperspectral image,'' \emph{IEEE Transactions on Geoscience and Remote Sensing}, vol.~59, no.~5, pp. 4191--4202, 2021.

\bibitem{uestc1}
Z.~Wen, Y.~Ling, Y.~Ren, T.~Wu, J.~Chen, X.~Pu, Z.~Hao, and L.~He, ``Homophily-related: Adaptive hybrid graph filter for multi-view graph clustering,'' \emph{arXiv preprint arXiv:2401.02682}, 2024.

\bibitem{uestc2}
C.~Cui, Y.~Ren, J.~Pu, J.~Li, X.~Pu, T.~Wu, Y.~Shi, and L.~He, ``A novel approach for effective multi-view clustering with information-theoretic perspective,'' \emph{Advances in Neural Information Processing Systems}, vol.~36, 2024.

\bibitem{uestc3}
C.~Cui, Y.~Ren, J.~Pu, X.~Pu, and L.~He, ``Deep multi-view subspace clustering with anchor graph,'' \emph{arXiv preprint arXiv:2305.06939}, 2023.

\bibitem{uestc4}
W.~Yan, Y.~Zhou, Y.~Wang, Q.~Zheng, and J.~Zhu, ``Multi-view semantic consistency based information bottleneck for clustering,'' \emph{Knowledge-Based Systems}, vol. 288, p. 111448, 2024.

\bibitem{uestc5}
Y.~Zhou, Q.~Zheng, Y.~Wang, W.~Yan, P.~Shi, and J.~Zhu, ``Mcoco: Multi-level consistency collaborative multi-view clustering,'' \emph{Expert Systems with Applications}, vol. 238, p. 121976, 2024.

\bibitem{15yang2020probabilistic}
L.~Yang, Y.~Guo, J.~Gu, D.~Jin, B.~Yang, and X.~Cao, ``Probabilistic graph convolutional network via topology-constrained latent space model,'' \emph{IEEE Transactions on Cybernetics}, vol.~52, no.~4, pp. 2123--2136, 2020.

\bibitem{wang2022graph}
J.~Wang, C.~Tang, X.~Zheng, X.~Liu, W.~Zhang, and E.~Zhu, ``Graph regularized spatial--spectral subspace clustering for hyperspectral band selection,'' \emph{Neural Networks}, vol. 153, pp. 292--302, 2022.

\bibitem{17cai2021graph}
Y.~Cai, M.~Zeng, Z.~Cai, X.~Liu, and Z.~Zhang, ``Graph regularized residual subspace clustering network for hyperspectral image clustering,'' \emph{Information Sciences}, vol. 578, pp. 85--101, 2021.

\bibitem{38Liu}
S.~Liu and H.~Wang, ``Graph convolutional optimal transport for hyperspectral image spectral clustering,'' \emph{IEEE Transactions on Geoscience and Remote Sensing}, vol.~60, pp. 1--13, 2022.

\bibitem{hao1}
Z.~Hao, Z.~Lu, G.~Li, F.~Nie, R.~Wang, and X.~Li, ``Ensemble clustering with attentional representation,'' \emph{IEEE Transactions on Knowledge \& Data Engineering}, vol.~36, no.~02, pp. 581--593, 2024.

\bibitem{hao2}
Z.~Hao, Z.~Lu, F.~Nie, R.~Wang, and X.~Li, ``Multi-view k-means with laplacian embedding,'' in \emph{ICASSP 2023-2023 IEEE International Conference on Acoustics, Speech and Signal Processing (ICASSP)}, 2023, pp. 1--5.

\bibitem{PSCPC}
R.~Guan, Z.~Li, X.~Li, and C.~Tang, ``Pixel-superpixel contrastive learning and pseudo-label correction for hyperspectral image clustering,'' in \emph{ICASSP 2024 - 2024 IEEE International Conference on Acoustics, Speech and Signal Processing (ICASSP)}, 2024, pp. 6795--6799.

\bibitem{NCSC}
Y.~Cai, Z.~Zhang, P.~Ghamisi, Y.~Ding, X.~Liu, Z.~Cai, and R.~Gloaguen, ``Superpixel contracted neighborhood contrastive subspace clustering network for hyperspectral images,'' \emph{IEEE Transactions on Geoscience and Remote Sensing}, vol.~60, pp. 1--13, 2022.

\bibitem{li2023mixture}
X.~Li, Z.~Ni, and T.~Zhang, ``Mixture of personality improved spiking actor network for efficient multi-agent cooperation,'' \emph{Frontiers in Neuroscience}, vol.~17, p. 1219405, 2023.

\bibitem{CMSCGC}
R.~Guan, Z.~Li, W.~Tu, J.~Wang, Y.~Liu, X.~Li, C.~Tang, and R.~Feng, ``Contrastive multi-view subspace clustering of hyperspectral images based on graph convolutional networks,'' \emph{IEEE Transactions on Geoscience and Remote Sensing}, vol.~62, pp. 1--14, 2024.

\bibitem{22liu2022multilayer}
L.~Liu, Z.~Kang, J.~Ruan, and X.~He, ``Multilayer graph contrastive clustering network,'' \emph{Information Sciences}, vol. 613, pp. 256--267, 2022.

\bibitem{24li2022deep}
T.~Li, Y.~Cai, Y.~Zhang, Z.~Cai, and X.~Liu, ``Deep mutual information subspace clustering network for hyperspectral images,'' \emph{IEEE Geoscience and Remote Sensing Letters}, vol.~19, pp. 1--5, 2022.

\bibitem{hutao1}
Q.~Yan, T.~Hu, Y.~Sun, H.~Tang, Y.~Zhu, W.~Dong, L.~Van~Gool, and Y.~Zhang, ``Towards high-quality hdr deghosting with conditional diffusion models,'' \emph{IEEE Transactions on Circuits and Systems for Video Technology}, 2023.

\bibitem{hutao2}
B.~Wang, T.~Hu, B.~Li, X.~Chen, and Z.~Zhang, ``Gatector: A unified framework for gaze object prediction,'' in \emph{Proceedings of the IEEE/CVF Conference on Computer Vision and Pattern Recognition}, 2022, pp. 19\,588--19\,597.

\bibitem{hutao3}
X.~Zhang, T.~Hu, J.~He, and Q.~Yan, ``Efficient content reconstruction for high dynamic range imaging,'' in \emph{ICASSP 2024 - 2024 IEEE International Conference on Acoustics, Speech and Signal Processing (ICASSP)}, 2024, pp. 7660--7664.

\bibitem{25brbic2018multi}
M.~Brbi{\'c} and I.~Kopriva, ``Multi-view low-rank sparse subspace clustering,'' \emph{Pattern Recognition}, vol.~73, pp. 247--258, 2018.

\bibitem{xia1}
Y.~Xia, X.~Shao, T.~Ding, and J.~Liu, ``Prescribed intelligent elliptical pursuing by uavs: A reinforcement learning policy,'' \emph{Expert Systems with Applications}, vol. 249, p. 123547, 2024.

\bibitem{xia2}
Z.~Wang, J.~Du, C.~Jiang, Z.~Zhang, Y.~Ren, and Z.~Han, ``Dynamic packet routing based on acoustic signal curve propagation in the auv-assisted iout,'' \emph{IEEE Internet of Things Journal}, vol.~11, no.~6, pp. 9854--9869, 2024.

\bibitem{xia3}
Z.~Wang, Z.~Zhang, J.~Wang, C.~Jiang, W.~Wei, and Y.~Ren, ``Auv-assisted node repair for iout relying on multiagent reinforcement learning,'' \emph{IEEE Internet of Things Journal}, vol.~11, no.~3, pp. 4139--4151, 2024.

\bibitem{xia4}
X.~Hou, J.~Wang, C.~Jiang, X.~Zhang, Y.~Ren, and M.~Debbah, ``Uav-enabled covert federated learning,'' \emph{IEEE Transactions on Wireless Communications}, vol.~22, no.~10, pp. 6793--6809, 2023.

\bibitem{xia5}
X.~Hou, J.~Wang, C.~Jiang, Z.~Meng, J.~Chen, and Y.~Ren, ``Efficient federated learning for metaverse via dynamic user selection, gradient quantization and resource allocation,'' \emph{IEEE Journal on Selected Areas in Communications}, vol.~42, no.~4, pp. 850--866, 2024.

\bibitem{hao3}
F.~Nie, Z.~Hao, and R.~Wang, ``Multi-class support vector machine with maximizing minimum margin,'' \emph{arXiv preprint arXiv:2312.06578}, 2023.

\bibitem{SCGN}
Y.~Liu, X.~Yang, S.~Zhou, X.~Liu, S.~Wang, K.~Liang, W.~Tu, and L.~Li, ``Simple contrastive graph clustering,'' \emph{IEEE Transactions on Neural Networks and Learning Systems}, pp. 1--12, 2023.

\bibitem{Dinknet}
Y.~Liu, K.~Liang, J.~Xia, S.~Zhou, X.~Yang, X.~Liu, and S.~Z. Li, ``Dink-net: Neural clustering on large graphs,'' in \emph{International Conference on Machine Learning}.\hskip 1em plus 0.5em minus 0.4em\relax PMLR, 2023, pp. 21\,794--21\,812.

\bibitem{DFCN}
W.~Tu, S.~Zhou, X.~Liu, X.~Guo, Z.~Cai, E.~Zhu, and J.~Cheng, ``Deep fusion clustering network,'' in \emph{Proceedings of the AAAI Conference on Artificial Intelligence}, vol.~35, no.~11, 2021, pp. 9978--9987.

\bibitem{DGCN}
Y.~Liu, W.~Tu, S.~Zhou, X.~Liu, L.~Song, X.~Yang, and E.~Zhu, ``Deep graph clustering via dual correlation reduction,'' in \emph{Proceedings of the AAAI conference on artificial intelligence}, vol.~36, no.~7, 2022, pp. 7603--7611.

\bibitem{HSAN}
Y.~Liu, X.~Yang, S.~Zhou, X.~Liu, Z.~Wang, K.~Liang, W.~Tu, L.~Li, J.~Duan, and C.~Chen, ``Hard sample aware network for contrastive deep graph clustering,'' in \emph{Proceedings of the AAAI conference on artificial intelligence}, vol.~37, no.~7, 2023, pp. 8914--8922.

\bibitem{44}
H.~Li, Y.~Li, G.~Zhang, R.~Liu, H.~Huang, Q.~Zhu, and C.~Tao, ``Global and local contrastive self-supervised learning for semantic segmentation of hr remote sensing images,'' \emph{IEEE Transactions on Geoscience and Remote Sensing}, vol.~60, pp. 1--14, 2022.

\bibitem{PCA}
C.~Rodarmel and J.~Shan, ``Principal component analysis for hyperspectral image classification,'' \emph{Surveying and Land Information Science}, vol.~62, no.~2, pp. 115--122, 2002.

\bibitem{33kanungo2002efficient}
T.~Kanungo, D.~M. Mount, N.~S. Netanyahu, C.~D. Piatko, R.~Silverman, and A.~Y. Wu, ``An efficient k-means clustering algorithm: Analysis and implementation,'' \emph{IEEE transactions on pattern analysis and machine intelligence}, vol.~24, no.~7, pp. 881--892, 2002.

\bibitem{8zhang2016spectral}
H.~Zhang, H.~Zhai, L.~Zhang, and P.~Li, ``Spectral--spatial sparse subspace clustering for hyperspectral remote sensing images,'' \emph{IEEE Transactions on Geoscience and Remote Sensing}, vol.~54, no.~6, pp. 3672--3684, 2016.

\bibitem{36zhang2019hyperspectral}
L.~Zhang, L.~Zhang, B.~Du, J.~You, and D.~Tao, ``Hyperspectral image unsupervised classification by robust manifold matrix factorization,'' \emph{Information Sciences}, vol. 485, pp. 154--169, 2019.

\end{thebibliography}

%
%

\end{document}